\documentclass[runningheads]{llncs}
\usepackage[utf8]{inputenc}
\usepackage{graphicx}
\usepackage{xcolor}
\usepackage{xspace}
\usepackage{listings}
\usepackage[title]{appendix}
\begin{document}
\title{A Preliminary Case Study of Planning With Complex Transitions: Plotting\thanks{
This work is supported by UK EPSRC EP/P015638/1 and EP/V027182/1, by the MICINN/FEDER, UE (RTI2018-095609-B-I00), by the French Agence Nationale de la Recherche, reference ANR-19-CHIA-0013-01, and by Archimedes institute, Aix-Marseille University.
}}
%
%
\author{
Jordi Coll\inst{1}\and
Joan Espasa\inst{2}\and
Ian Miguel\inst{2}\and
Mateu Villaret\inst{3}
}
\authorrunning{J. Coll et al.}
%
\institute{Princeton University, Princeton NJ 08544, USA \and
Springer Heidelberg, Tiergartenstr. 17, 69121 Heidelberg, Germany
\email{lncs@springer.com}\\
\url{http://www.springer.com/gp/computer-science/lncs} \and
ABC Institute, Rupert-Karls-University Heidelberg, Heidelberg, Germany\\
\email{\{abc,lncs\}@uni-heidelberg.de}}

\institute{
    Aix Marseille Univ, Universit\'e de Toulon, CNRS, LIS, Marseille, France \\
    \email{jordi.coll@lis-lab.fr} \and
    School of Computer Science, University of St Andrews, St Andrews KY16 9SX, UK
    \email{\{jea20,ijm\}@st-andrews.ac.uk} \and
    Departament d'Inform\`atica, Matem\`atica Aplicada i Estad\'istica \\
    Universitat de Girona, E-17003 Girona, Spain \\
    \email{mateu.villaret@udg.edu}
}

\maketitle              

\newcommand{\eprime}{{Essence Prime}\xspace}
\newcommand{\savilerow}{{Savile Row}\xspace}
\newcommand{\joan}[1]{\colorbox{orange}{#1}}
\newcommand{\comment}[1]{{\color{red}#1}}
\newcommand{\define}{=}
\newcommand{\Pre}{\mathit{Pre}}
\newcommand{\Eff}{\mathit{Eff}}
\newcommand{\true}{\mathit{true}}
\newcommand{\false}{\mathit{false}}

\begin{abstract}
Plotting is a tile-matching puzzle video game published by Taito in 1989. Its objective is to reduce a given grid of coloured blocks down to a goal number or fewer. This is achieved by the avatar character repeatedly shooting the block it holds into the grid. Plotting is an example of a planning problem: given a model of the environment, a planning problem asks us to find a sequence of actions that can lead from an initial state of the environment to a given goal state while respecting some constraints. The key difficulty in modelling Plotting is in capturing the way the puzzle state changes after each shot. A single shot can affect multiple tiles directly, and the grid is affected by gravity so numerous other tiles can be affected indirectly. We present and evaluate a constraint model of the Plotting problem that captures this complexity. We also discuss the difficulties and inefficiencies of modelling Plotting in PDDL, the standard language used for input to specialised AI planners. We conclude by arguing that AI planning could benefit from a richer modelling language.

\end{abstract}
%
%
%
\section{Introduction}

\begin{figure}[t!] \label{fig:screenshot}
\centering
\includegraphics[scale=0.3]{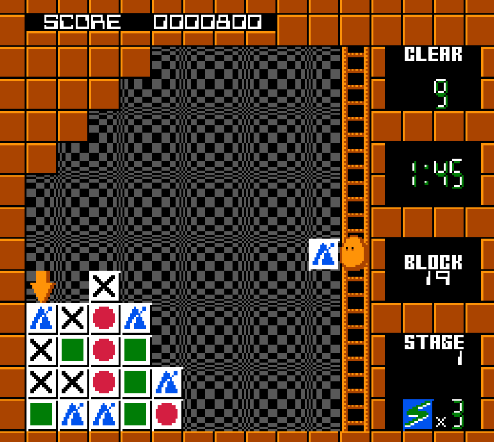}
\caption{Screen shot from the Plotting video game (Taito, 1989).}
\end{figure}

Plotting is a puzzle video game published by Taito in 1989 and ported to many platforms. The objective of the game is to reduce a given grid of coloured blocks to a goal number or fewer (Figure \ref{fig:screenshot}). This is achieved by the avatar character repeatedly shooting the block it holds into the grid. It is called \emph{Flipull} in Japan as well as in versions for the Famicom and Game Boy, and \emph{Plotting} elsewhere.

Plotting is an example of a planning problem: given a model of the environment, a planning problem asks us to find a sequence of actions that can lead from an initial state of the environment to a given goal state while respecting some constraints. Planning is an essential part of decision making, and therefore a central discipline of Artificial Intelligence (AI). Examples of its usage in industry and academia are many and varied, such as logistics~\cite{usecase3}, chemistry~\cite{usecase2} or drilling operations~\cite{usecase1}.

In Plotting, we are trying to find a sequence of positions to fire from such that enough blocks are removed to beat the current scenario. The game is played by one agent, which has the full information of the state and the effects of each action are deterministic. This situation helps into finding approaches to model and solve the problem, and its similar to pen and paper puzzle games~\cite{puzzles} or board games such as peg solitaire~\cite{pegsolitaire} or the knight's tour~\cite{knightstour}.

The Planning Domain Definition Language (PDDL)~\cite{pddl} is the de-facto standard modelling language for planning problems, supported by
most planning systems. Its widespread use started thanks to the collaborative efforts and desire of the community to facilitate benchmarking 
and applications of planning systems. When using PDDL, the user describes the problem in terms of predicates, 
actions and functions with parameters. In turn, these parameters are instantiated with a set of defined objects.

In this work we will show that modelling the complex dynamics of the game in PDDL is not straightforward. The resulting complexity of the model
hinders the efficacy of planning systems to produce a valid plan. We also propose a model of the game in 
\eprime\cite{essence-prime-description}, a declarative constraint modelling language able to express planning problems~\cite{modref20}.
Moreover, we take advantage of \savilerow~\cite{savilerow}, a sophisticated constraint reformulation tool supporting \eprime that is able 
to generate SAT and SMT~\cite{biere2021handbook}, or CSP~\cite{rossi2006handbook} instances. 

Our contributions in this paper are: models for Plotting in \eprime and PDDL,
a parameterised instance generator, and an empirical evaluation of the proposed \eprime model.
 
\section{Background}


A \emph{classical planning problem} can be defined as a tuple
$\prod = \langle F,A,I,G\rangle$, where: $F$ is a set of propositional state
variables,  $A$ is a set of actions, $I$ is the initial state and $G$ is the goal.
Given a planning problem $\prod$, a \emph{state} is a variable-assignment (or
valuation) function over state variables $F$, which maps each
variable of $F$ into a truth value.

An action $a \in A$ is defined as a tuple $a = \langle \Pre_a, \Eff_a \rangle$,
where $\Pre_a$ refers to the preconditions and $\Eff_a$ to the effects of the
action.
Preconditions ($\Pre$) and the goal $G$ are first-order formulas  over propositional
state variables. Action effects ($\Eff$) are sets of assignments to propositional state variables.

An action $a$ is \emph{applicable} in a state $s$ only if its
precondition is satisfied in $s$ ($s\models\Pre_a$). The outcome after the application of an
action $a$ will be the state where variables that are assigned in
$\Eff_a$ take their new value, and variables not referenced in
$\Eff_a$ keep their current values.
A sequence of actions $\langle a_0, \dots, a_{n-1} \rangle$ is called a
\emph{plan}. We say that the application of a plan starting from the initial
state $I$ brings the system to a state $s_n$. If each action is applicable in
the state resulting from the application of the previous action and the final
state satisfies the goal (i.e., $s_n \models G$), the sequence of actions is a
a \emph{valid plan}. A planning problem has a solution if a valid plan can be found for the problem.

\subsection{Planning as Satisfiability}
\label{sec:PlanningAsSatisfiability}

In contrast with games such as peg solitaire~\cite{pegsolitaire}, we cannot a priori compute the length of the plan. When encoding a planning problem into CP or SAT, it is common in this situation~\cite{KautzS92,cplan,miguel2000flexible} to solve the planning problem by considering a sequence of satisfaction problems $\phi_0$, $\phi_1$, $\phi_2$, \dots, where $\phi_i$ encodes the existence of a plan that reaches a goal state from the initial state in $i$ steps.



In constructing each $\phi$, we take the common approach \cite{pegsolitaire,gent2007search} of formulating a ``state and action'' constraint model of the planning problem, where we employ decision variables to capture both the state of the puzzle at each time step and the action taken to transform the preceding into the succeeding state. Constraints ensure that when an action is executed, its preconditions hold with respect to the problem variables and that its effects are applied and the state is modified accordingly. Constraints on the variables representing the state of the final step require that the goal conditions are met. Finally, \textit{frame axioms} are made explicit, i.e. constraints that specify that if no action has modified a variable, it keeps its value between steps. Semantics such as the $\forall$ or $\exists$-step~\cite{RintanenHN06} allow parallel actions, but here just one action will be executed per time step. 

\section{Plotting}

The objective in Plotting is to reduce a given grid of coloured blocks down to a goal number or fewer. This is achieved by the avatar character shooting the block it holds into the grid, either horizontally directly into the grid, or by shooting at the wall blocks above the grid, and bouncing down vertically onto the grid. The rules for what happens when a block hits the wall are as follows:
\begin{itemize}
\item  If it hits a wall as it is travelling horizontally, it falls vertically downwards. Note that in some levels, additional walls are arranged to facilitate hitting the blocks from above.
This arrangement varies with instances of the puzzle — in harder instances wall cells are placed so as to prevent throwing blocks along some rows and columns.
\item  If it falls onto a wall, it rebounds into the avatar’s hand.
\end{itemize}

The rules for a shot block $S$ colliding with a block $B$ in the grid:
\begin{itemize}
\item  If the first block $S$ hits is of a different type from itself, $S$ rebounds into the avatar’s hand and the grid is unchanged --- a null move.
    \item  If \emph{S} and \emph{B} are of the same type, \emph{B} is consumed and \emph{S} continues to travel in the same direction. All blocks above \emph{B} fall one grid cell each.
    \item  If \emph{S}, having already consumed a block of the same type, hits a block \emph{B} of a different type, \emph{S} replaces \emph{B}, and \emph{B} rebounds into the avatar’s hand.
\end{itemize}

The combination of the second and third rules above means that it is possible to shoot through an entire row of the grid, hit the wall, and continue to consume blocks as the shot block falls. If, after making a shot, the block that rebounds into the avatar’s hand is such that there is now no possible shot that can further reduce the grid, the player loses a life and the block in the avatar’s hand is transformed into a wildcard block, which transforms into the same type as the first block it hits. Each level also begins with the avatar holding a wildcard block.

When considered as a planning problem, Plotting's initial state is the given grid, and there are usually multiple goal states where the grid is sufficiently reduced to meet the target. We abstract out the avatar's movement to consider the key decisions: the rows or columns chosen at which to shoot the held blocks. Therefore, the sequence of actions to get us from the initial to the goal state is comprised of individual shots at the grid, either horizontally or vertically.

\section{Constraint Model in Essence Prime}\label{sec:eprime_model}

Rendl et al.~\cite{sara09} provide a brief description of an incomplete constraint model of Plotting. We give here a full description of a complete model of the problem, formulated in a state and action style, as noted in Section \ref{sec:PlanningAsSatisfiability}. Here, the state is the current grid configuration and the contents of the hand of the avatar, and the single action is a shot along a particular row or column.



\subsection{Preliminaries: Constants and Parameters}


Each block type is identified with a colour, and the colours are represented by a contiguous range of natural numbers in Essence Prime. We use 0 to represent an empty grid cell. Step 0 will represent the initial state, with the action chosen at step 1 transforming the initial state into the state at step 1, and so on. The parameters and constants for the model are therefore:

\begin{scriptsize}
\begin{verbatim}
    given initGrid : matrix indexed by[int(1..gridHeight), int(1..gridWidth)] of int(1..)
    letting GRIDCOLS be domain int(1..gridWidth)
    letting GRIDROWS be domain int(1..gridHeight)
    letting NOBLOCKS be gridWidth * gridHeight
    letting COLOURS be domain int(1..max(flatten(initGrid)))
    letting EMPTY be 0
    letting EMPTYANDCOLOURS be domain int(EMPTY) union COLOURS

    given goalBlocksRemaining : int(1..NOBLOCKS)
    given noSteps : int(1..)
    letting STEPSFROM1 be domain int(1..noSteps)
    letting STEPSFROM0 be domain int(0..noSteps) 
\end{verbatim}
\end{scriptsize}

\subsection{Basic Viewpoint}

We capture the current state of the grid and the contents of the avatar's hand at each time step with a time-indexed 2d array of decision variables and an individual variable per time step respectively. Only one action is possible per time step, which is the decision as to where to fire the block held. Here we introduce a pair of variables per time step, one representing the column fired down (if any) and one representing the row fired along (if any):

\begin{scriptsize}
\begin{verbatim}
    find fpRow : matrix indexed by[STEPSFROM1] of int(0..gridHeight)
    find fpCol : matrix indexed by[STEPSFROM1] of int(0..gridWidth)
    find grid : matrix indexed by[STEPSFROM0, GRIDROWS, GRIDCOLS]
                               of EMPTYANDCOLOURS 
    find hand : matrix indexed by[STEPSFROM0] of COLOURS
\end{verbatim}
\end{scriptsize}

There is also a set of auxiliary variable to support the special case of when a block consumes an entire row and then falls through more blocks on the rightmost column. We will return to these later.

\subsection{Initial, Goal States}

The initial state is as simple as constraining the 0th 2d array of {\tt grid} to be equal to the parameter {\tt initGrid}. The goal state is expressed simply by counting the number of empty grid cells:
\begin{scriptsize}
\begin{verbatim}
    $ Initial state:
    forAll gCol : GRIDCOLS .
      forAll gRow : GRIDROWS .
        grid[0, gRow, gCol] = initGrid[gRow, gCol],
    $ Goal state:
    atleast(flatten(grid[noSteps,..,..]),
            [NOBLOCKS - goalBlocksRemaining],
            [EMPTY]),
\end{verbatim}
\end{scriptsize}

\subsection{Constraining the Actions}

Having transformed Plotting into a decision problem that asks if there is a plan for with a fixed number of steps, we might take the view that moves that don't alter the state of the puzzle (e.g. firing the held block straight into one of a different colour) might be used to ``pad'' a short plan to the given length. We decided that this was of little benefit and otherwise could lead to redundant search, and so  disallowed moves that don't progress the solution of the puzzle with the following sum constraint:

\begin{scriptsize}
\begin{verbatim}
    $ Each move must do something useful:
    forAll step : STEPSFROM1 .
      sum(flatten(grid[step-1,..,..])) > sum(flatten(grid[step,..,..])),
\end{verbatim}
\end{scriptsize}

This does mean care will be necessary with our frame constraints, as we discuss in Section \ref{sec:ConstrainingState} below. Any cell unconstrained will be vulnerable to the constraint solver assigning an arbitrary (low-numbered) colour so as to satisfy the sum constraint above.

The other constraint we consider here is to say that we must fire horizontally or vertically but not both:

\begin{scriptsize}
\begin{verbatim}
    $ Exactly one fp axis must be 0. (exclusive OR, only ONE fired angle)
    forAll step : STEPSFROM1 .
      (fpRow[step] * fpCol[step]) = 0 /\ (fpRow[step] + fpCol[step]) > 0,
\end{verbatim}
\end{scriptsize}
      
\subsection{Constraining State}
\label{sec:ConstrainingState}

There remain the constraints specifying valid state changes. These are all stated in an if-and-only-if form to ensure that no part of the state (hand or grid) is left unconstrained and therefore vulnerable to the solver assigning arbitrary values. These constraints are subdivided into four cases:
\begin{small}
\begin{enumerate}
    \item  The hand is unchanged.
    \item  A grid cell becomes empty.
    \item  A grid cell stays the same.
    \item  A grid cell changes colour to something other than empty. This can affect the hand.
\end{enumerate}
\end{small}

Some of these interact with the special case that was missing from the Rendl et al. model (i.e. a block is fired horizontally, hits the wall and falls, continuing to consume blocks). The references to ``wallFall'', which is explained in the next section, are related to this. The wallFall value records how many cells a block undermined by the special case situation will fall in the last column.

There are two cases in which the hand is unchanged when we require a progressing move. The first is firing down a column containing only the same colour blocks as the block fired. The second is along a row of the same colour, hitting the wall, then consuming everything beneath it on that rightmost column before hitting the floor:

\begin{scriptsize}
\begin{verbatim}
    forAll step : STEPSFROM1 .
      (hand[step-1] = hand[step])
      =
      (
        $ Fired down col, hitting wall
        (
          (forAll colBlock : GRIDROWS .
            ((grid[step-1,colBlock,fpCol[step]] = hand[step-1]) \/
             (grid[step-1,colBlock,fpCol[step]] = EMPTY))
          )
        ) \/
        $ Fired row, hitting wall, dropping through hand-colour only.
        (
          $ along the row
          (
            forAll rowBlock : GRIDCOLS .
              ((grid[step-1,fpRow[step],rowBlock] = hand[step-1]) \/
               (grid[step-1,fpRow[step],rowBlock] = EMPTY))
          ) /\
          $ down the column
          (
           forall rowBeneath : GRIDROWS .
             (rowBeneath > fpRow[step] ->
               ((grid[step-1,rowBeneath,gridWidth] = hand[step-1]) \/
                (grid[step-1,rowBeneath,gridWidth] = EMPTY))
             ) ) ) ),
\end{verbatim}
\end{scriptsize}

There are six cases governing how a grid cell becomes empty, including the trivial case of it being empty at the previous time step:
\begin{scriptsize}
\begin{verbatim}
    forAll step : STEPSFROM1 .
      forAll gRow : GRIDROWS .
        forAll gCol : GRIDCOLS .
          (grid[step,gRow,gCol] = EMPTY)
          =
          ( $ When a cell is EMPTY, it stays EMPTY
            (grid[step-1,gRow,gCol] = EMPTY) \/
            $ Deleted by shot down column
            (
              $ The right column
              (fpCol[step] = gCol) /\
              $ same colour as hand
              (grid[step-1,gRow,gCol] = hand[step-1]) /\
              $ Nothing blocking the way
              (forAll blockAbove : int(1..gRow-1) .
                ((grid[step-1,blockAbove,fpCol[step]] = hand[step-1]) \/
                 (grid[step-1,blockAbove,fpCol[step]] = EMPTY))
              )
            ) \/ $ Deleted by shot along row
            (
              $ The right row
              (fpRow[step] = gRow) /\
              $ same colour as hand
              (grid[step-1,gRow,gCol] = hand[step-1]) /\
              $ no block above
              ((gRow = 1) \/
               (grid[step-1,gRow-1,gCol] = EMPTY)) /\
              $ nothing blocking way
              (forAll blockLeft : int(1..gCol-1) .
                ((grid[step-1,gRow,blockLeft] = hand[step-1]) \/
                 (grid[step-1,gRow,blockLeft] = EMPTY))
              )
            ) \/ $ Deleted by shot along row, then down col
            (
              $ Only the final column
              (gCol = gridWidth) /\
              $ fpRow is above this
              (fpRow[step] < gRow) /\
              $ same colour as hand
              (grid[step-1,gRow,gridWidth] = hand[step-1]) /\
              $ Nothing blocking the way on the row
              (forAll rowBlock : int(1..gridWidth) .
                 ((grid[step-1,fpRow[step],rowBlock] = hand[step-1]) \/
                  (grid[step-1,fpRow[step],rowBlock] = EMPTY))
              ) /\
              $ Nothing blocking the way on the final column
              (forAll colBlock : int(1..gRow-1) .
                 (colBlock > fpRow[step]) ->
                 ((grid[step-1,colBlock,gridWidth] = hand[step-1]) \/
                  (grid[step-1,colBlock,gridWidth] = EMPTY))
              ) /\
              $ Empty above the firing row on the final column
              (forAll colBlock : int(1..gRow-1) .
                 (colBlock < fpRow[step]) ->
                 (grid[step-1,colBlock,gridWidth] = EMPTY)
              )
            ) \/ $ Fall from this cell to become empty - row shot underneath
            (
              $ There was no block above
              ((grid[step-1,gRow-1,gCol] = EMPTY) \/
               (gRow = 1)) /\
              $ Deletion below
              (fpRow[step] > gRow) /\
              (forAll delBlock : int(1..gCol) .
                ((grid[step-1,fpRow[step],delBlock] = hand[step-1]) \/
                 (grid[step-1,fpRow[step],delBlock] = EMPTY)))
            ) \/
            $ Final Column shot along a row consuming several blocks underneath
            (
              $ Only the final column
              (gCol = gridWidth) /\
              $ There was a wallfall - this implies a successful row shot.
              (wallFall[step] > 0) /\
              $ The shot was beneath here
              (fpRow[step] > gRow) /\
              $ Nothing there to fall into here
              (grid[step-1,gRow-wallFall[step],gridWidth] = EMPTY \/
               gRow-wallFall[step] < 1)
            ) ),
\end{verbatim}
\end{scriptsize}
          
There are 9 cases governing how a grid cell can remain the same. The test is simple equality, so we have to include emptiness here:
\begin{scriptsize}
\begin{verbatim}
    forAll step : STEPSFROM1 .
      forAll gRow : GRIDROWS .
        forAll gCol : GRIDCOLS .
          (grid[step,gRow,gCol] = grid[step-1,gRow,gCol])
          =
          ( $ It was empty
            (grid[step-1,gRow,gCol] = EMPTY) \/
            $ Fired beneath this row, not far enough to cause fall:
            (
              (fpRow[step] > gRow) /\
              (exists blockLeft : int(1..gCol) .
                 ((grid[step-1,fpRow[step],blockLeft] != EMPTY) /\
                  (grid[step-1,fpRow[step],blockLeft] != hand[step-1]))
              )
            ) \/ $ Fired along this row, but something in the way
            (
              (fpRow[step] = gRow) /\
              (exists blockLeft : int(1..gCol-1) .
                ((grid[step-1, gRow, blockLeft] != EMPTY) /\
                 (grid[step-1, gRow, blockLeft] != hand[step-1]))
              )
            ) \/ $ Fired along row above, cols except last
            (
              (gCol < gridWidth) /\
              (fpRow[step] != 0) /\
              (fpRow[step] < gRow)
            ) \/
            $ Fired along row above, last col. Sth in way on row or last col.
            (
              (gCol = gridWidth) /\
              (fpRow[step] != 0) /\
              (fpRow[step] < gRow) /\
              (
                (exists rowBlock : int(1..gridWidth) .
                 ((grid[step-1, fpRow[step], rowBlock] != EMPTY) /\
                  (grid[step-1, fpRow[step], rowBlock] != hand[step-1]))
                ) \/
                (exists colBlock : int(1..gRow-1) .
                 ((colBlock >= fpRow[step]) /\
                  (grid[step-1, colBlock, gridWidth] != EMPTY) /\
                  (grid[step-1, colBlock, gridWidth] != hand[step-1]))
                )
              )
            ) \/
            $ Fired down this column, but something in way
            (
              (fpCol[step] = gCol) /\
              (exists blockAbove : int(1..gRow-1) .
                ((grid[step-1,blockAbove,gCol] != EMPTY) /\
                 (grid[step-1,blockAbove,gCol] != hand[step-1])))
            ) \/
            $ Fired down a different column
            (
              (fpCol[step] != 0) /\
              (fpCol[step] != gCol)
            ) \/
            $ This row or below. Same colour block falls here. All but last col.
            (
              (gCol < gridWidth) /\
              (fpRow[step] >= gRow) /\
              (forAll delBlock : int(1..gCol) .
                ((grid[step-1,fpRow[step],delBlock] = hand[step-1]) \/
                 (grid[step-1,fpRow[step],delBlock] = EMPTY))) /\
              (grid[step-1,gRow-1,gCol] = grid[step-1,gRow,gCol])
            ) \/
            $ This row or below. Same colour block falls here. Last col.
            (
              (gCol = gridWidth) /\
              (fpRow[step] >= gRow) /\
              (wallFall[step] > 0) /\
              (grid[step-1,gRow-wallFall[step],gCol] = grid[step-1,gRow,gCol])
            )
          ),
\end{verbatim}
\end{scriptsize}
          
Finally, five cases for a grid cell changing to something other than empty:
\begin{scriptsize}
\begin{verbatim}
    forAll step : STEPSFROM1 .
      forAll gRow : GRIDROWS .
        forAll gCol : GRIDCOLS .
          ((grid[step,gRow,gCol] != grid[step-1,gRow,gCol]) /\
           (grid[step,gRow,gCol] != EMPTY))
          =
          (
            $ Fall from above. Not rightmost col.
            (
              (gCol < gridWidth) /\
              $there was a block above
              (grid[step-1,gRow-1,gCol] != EMPTY) /\
              $Deletion here or below
              (fpRow[step] >= gRow) /\
              (forAll delBlock : int(1..gCol) .
                ((grid[step-1,fpRow[step],delBlock] = hand[step-1]) \/
                 (grid[step-1,fpRow[step],delBlock] = EMPTY))) /\
              $ Is now the same as the block above.
              (grid[step,gRow,gCol] = grid[step-1,gRow-1,gCol]) /\
              $ Which was a different colour
              (grid[step-1,gRow,gCol] != grid[step-1,gRow-1,gCol])
            ) \/
            $ Fall from above. Rightmost col.
            (
              (gCol = gridWidth) /\
              $ WallFall implies successful row shot
              (wallFall[step] > 0) /\
              $ Shot here or below
              (fpRow[step] >= gRow) /\
              $ Is now the same as the block above
              (grid[step,gRow,gridWidth] =
               grid[step-1,gRow-wallFall[step],gridWidth]) /\
              $ Which was a different colour
              (grid[step-1,gRow,gridWidth] !=
               grid[step-1,gRow-wallFall[step],gridWidth])
            ) \/
            $ Cell swaps with hand: row shot.
            (
              (gRow = fpRow[step]) /\
              $ The row shot
              (forAll colsLeft : int(1..gCol-1) .
                (grid[step-1,fpRow[step],colsLeft] = hand[step-1]) \/
                (grid[step-1,fpRow[step],colsLeft] = EMPTY)) /\
              $ At least one cell has to match the hand
              (exists colsLeft : int(1..gCol-1) .
                (grid[step-1,fpRow[step],colsLeft] = hand[step-1])) /\
              $ Exchanges with the hand
              (hand[step] = grid[step-1,fpRow[step],gCol]) /\
              (hand[step-1] = grid[step,fpRow[step],gCol]) /\
              $ Which was a different colour
              (hand[step-1] != grid[step-1,fpRow[step],gCol])
            ) \/
            $ Cell swaps with hand: col shot
            (
              (gCol = fpCol[step]) /\
              $ The col shot
              (forAll rowsAbove : int(1..gRow-1) .
                (grid[step-1,rowsAbove,fpCol[step]] = hand[step-1]) \/
                (grid[step-1,rowsAbove,fpCol[step]] = EMPTY)) /\
              $ At least one cell has to match the hand
              (exists rowsAbove : int(1..gRow-1) .
                (grid[step-1,rowsAbove,fpCol[step]] = hand[step-1])) /\
              $ Exchanges with the hand
              (hand[step] = grid[step-1,gRow,fpCol[step]]) /\
              (hand[step-1] = grid[step,gRow,fpCol[step]]) /\
              $ Which was a different colour
              (hand[step-1] != grid[step-1,gRow,fpCol[step]])
            ) \/
            $ Cell swaps with hand: row then down last col.
            (
              $ rightmost col
              (gCol = gridWidth) /\
              $ fpRow is above this
              (fpRow[step] < gRow) /\
              $ Nothing blocking the way on the row
              (forAll rowBlock : int(1..gridWidth-1) .
                ((grid[step-1,fpRow[step],rowBlock] = hand[step-1]) \/
                 (grid[step-1,fpRow[step],rowBlock] = EMPTY))
              ) /\
              $ Nothing blocking the way on the final column
              (forAll colBlock : int(1..gRow-1) .
                (colBlock >= fpRow[step]) ->
                  ((grid[step-1,colBlock,gridWidth] = hand[step-1]) \/
                   (grid[step-1,colBlock,gridWidth] = EMPTY))
              ) /\
              $ But there must exist one hand block on firing row or final col.
              ((exists rowBlock : int(1..gridWidth-1) .
                  grid[step-1,fpRow[step],rowBlock] = hand[step-1])
               \/
               (exists colBlock : int(1..gRow-1) .
                  colBlock >= fpRow[step] /\
                  grid[step-1,colBlock,gridWidth] = hand[step-1])
              ) /\
              $ Exchanges with hand
              (hand[step] = grid[step-1,gRow,gridWidth]) /\
              (hand[step-1] = grid[step,gRow,gridWidth]) /\
              $ Which was a different colour
              (hand[step-1] != grid[step-1,gRow,gridWidth])
            )
          ),
\end{verbatim}
\end{scriptsize}
\subsection{WallFall}

As is evident from the main constraints in the previous section, we have introduced a set of variables (one per time step) to capture the distance that blocks in the last column fall as a consequence of a block being shot horizontally, reaching the wall, and then optionally consuming blocks as it falls down the last column:
\begin{small}
\begin{verbatim}
    find wallFall : matrix indexed by[STEPSFROM1] of int(0..gridHeight)
\end{verbatim}
\end{small}
    
The constraints to make the calculation enumerate each possible value for the {\tt wallFall} variable and stipulate what must be true for that value to be valid:

\begin{scriptsize}
\begin{verbatim}
    forAll step : STEPSFROM1 .
     forAll i : int (1..gridHeight) .
      (wallFall[step] = i)
      =
      (exists row : int(2..gridHeight) .
        (fpRow[step] = row) /\
        $ Consumed row to the rightmost column
        (forAll col : int(1..gridWidth) .
          grid[step-1,row,col] = EMPTY \/
          grid[step-1,row,col] = hand[step-1]) /\
        $ Something to fall
        (grid[step-1,row-1,gridWidth] != EMPTY) /\
        $ Fell this far
        (forAll underRow : int (row..row+i-1) .
          grid[step-1,underRow,gridWidth] = hand[step-1]) /\
        $ And no further
        ((grid[step-1,row+i,gridWidth] != hand[step-1]) \/
         (row+i > gridHeight))
      ),
\end{verbatim}
\end{scriptsize}

\section{Model in PDDL}

 
PDDL~\cite{pddl} is an expressive modelling language, able to encode many real-life problems with complex dynamics. In spite of that, the complexity of its many features resulted in most AI planners lagging behind, supporting only a small core set of features. 
Fast Downward~\cite{fastdownward} is the most well-known, supported and reused state-of-the-art planner. Its preprocessing module performs sophisticated transformations from PDDL to the more solver-amenable SAS+ format~\cite{saspus}. This preprocessor is currently used by many of the state-of-the-art planners.
Therefore, we focus on modeling the Plotting problem using a subset of features supported by Fast Downward. This implies that we cannot natively use numeric state variables, multi-valued variables nor function symbols. 

PDDL is \emph{action-oriented}, in the sense that a PDDL model defines  what are the possible actions to do at each step. Also for each action, we must define the \texttt{precondition} over the state of the previous time step required to perform the action, and what is the \texttt{effect} over the state when that action is done. This contrasts with the Essence Prime model given in Section~\ref{sec:eprime_model}, that we might say that is \emph{state-oriented}: we add a constraint for each possible change of state, requiring that this change is coherent with the state at the previous time step and the chosen action.

We have designed a PDDL model that can be found in Appendix~\ref{pddlmodel}. In this section we provide some pieces of the model in order to illustrate the drawbacks of using this fragment of PDDL.
The viewpoint uses two types of objects: \texttt{colour} and \texttt{number}. Note that \texttt{number} will be the name of a type that we will use to manually encode the basic required numerical properties. The predicate \texttt{hand} has one colour parameter and encodes if the avatar is holding the given colour. The  \texttt{coloured} predicate expresses, given a row, column and colour, if the block is coloured in the given colour.

{\scriptsize
\begin{verbatim}
        (hand ?c - colour)
        (coloured ?row ?col - number ?c - colour)
\end{verbatim}
}

A few auxiliary predicates such as \texttt{islastcolumn} or \texttt{isbottomrow} are added to
make the model easier to read and remove the usage of as many quantifiers as possible.

{\scriptsize
\begin{verbatim}
        (isfirstcolumn ?n - number)
        (islastcolumn ?n - number)
        (istoprow ?n - number)
        (isbottomrow ?n - number)
\end{verbatim}
}

In Plotting, an essential requirement is the use of numbers in counting and calculation. For example, it is required to be able to refer to a specific cell in the grid or to calculate how many positions a cell has to fall after removing part of a column. Although important, numeric planning is not widely supported. Therefore, we need to encode some integer relations as Boolean predicates:

{\scriptsize
\begin{verbatim}
    (succ ?p1 ?p2 - number)         ; p1 is successor of p2
    (lt ?p1 ?p2 - number)           ; p1 is less than p2
    (distance ?p1 ?p2 ?p3 - number) ; p3 is p2 - p1
\end{verbatim}
}

Those are the successor and $<$ operators and the distance between two numbers. Notice that, these predicates will have to be defined in each instance file, along the specific scenario information. For instance when dealing with a $5\times 5$ board we need to state {\tt succ} for every pair of successive numbers between 1 and 5, and {\tt lt} and {\tt distance} for every pair of two numbers $(p1,p2)$ between 1 and 5 such that $p1<p2$. 
In Figure~\ref{fig:pddl-code} we provide an excerpt of the action consisting of partially removing cells of color {\tt ?c} in row {\tt ?r} until column {\tt ?t}, i.e. not reaching the last column.

\lstset{
  basicstyle=\ttfamily,
  basicstyle=\scriptsize,
  columns=fullflexible,
  keepspaces=true,
  numbers=left,
  tabsize=2,
stepnumber=1
}
\begin{figure}[ht!]
\begin{lstlisting}[escapechar=|]
(:action shoot-partial-row
    ;; ?r - what row we are shooting at
    ;; ?t - to, the "limiting" cell
    ;; ?c - the colour of the range we are removing
    :parameters (?r - number ?t - number ?c - colour)
    :precondition
    (and
        ;; ?col is the successor of ?t with a different colour than ?c
        (exists (?col - number)
            (and  (succ ?col ?t) 
                  (not (coloured ?r ?col ?c))
                  (not (coloured ?r ?col null))))
        ...
        ;; all the blocks up to ?t have either the colour ?c or are null
        (forall (?col - number) 
            (or  (lt ?t ?col) 
                 (and (= ?col ?t) (coloured ?r ?t ?c))
                 (or 
                    (coloured ?r ?col ?c)
                    (coloured ?r ?col null)))))
    :effect
    (and
        ;; Change hands colour and
        ;; The next cell that we cannot remove gets the colour from the hand
        (forall (?nextcolumn - number ?nextcolour - colour)|\label{line:nextcolorquant}|
            (when |\label{line:when}|
                (and
                    (succ ?nextcolumn ?t)
                    (coloured ?r ?nextcolumn ?nextcolour))|\label{line:nextcolourwhen}|
                (and 
                    (not (coloured ?r ?nextcolumn ?nextcolour))
                    (coloured ?r ?nextcolumn ?c)
                    (hand ?nextcolour) |\label{line:hand1}|
                    (not (hand ?c))))) |\label{line:hand2}|
        ;; finally move everything downwards.
        ... )
)
\end{lstlisting}
\caption{Fragment of the action \emph{shoot-partial-row} of the the PDDL model.}
\label{fig:pddl-code}
\end{figure}

The lack of multi-valued variables in the considered fragment of PDDL 
makes the encoding of some transitions not ideal.
For example, when changing the colour held by the avatar we generally need to state: \emph{remove any previous colour in the hand  and set the new colour}. This is done in lines \ref{line:hand1}-\ref{line:hand2}. However, having multi-valued variables would make this change straightforward and less error-prone. 
Moreover, due to the lack of support for function symbols in the considered PDDL fragment, we are forced to add extra quantifiers to the model to be able to name specific objects. For instance, the column of the cell next to {\tt ?t} (\texttt{?nextcolumn}) and its colour (\texttt{?nextcolour}) have to be discovered in the effects of formula. This quantification is introduced in line \ref{line:nextcolorquant}, and the values of \texttt{?nextcolumn} and \texttt{?nextcolour} are discovered in lines~\ref{line:when}-\ref{line:nextcolourwhen} as a condition for the effect to take place.


Note that if we could use function symbols and arithmetic, we could remove variables \texttt{?nextcolumn} and \texttt{?nextcolour}. We would only have to change the \texttt{coloured} symbol to be a function that, given a row and column it would map to the colour in that cell.
Overall, lines \ref{line:nextcolorquant}-\ref{line:hand2} could theoretically be simplified as:

{\scriptsize
\begin{verbatim}
    (assign (hand (coloured ?r (?t + 1))))
    (assign (coloured ?r (?t + 1)) ?c)
\end{verbatim}
}

Note that the Essence Prime language naturally deals with this kind of statements, and in fact similar statements are used in the model given in Section~\ref{sec:eprime_model}. Assuming we have \texttt{hand} and \texttt{coloured} variables indexed by time step, the equivalent in Essence Prime would be:

{\scriptsize
\begin{verbatim}
    hand[step]=coloured[step-1][r][t+1]   /\   coloured[step][r][t+1]=c
\end{verbatim}
}

Finally, we will need to define the initial and goal states for every instance. The initial state can be straightforwardly stated with a  \texttt{coloured} statement for each cell. However, the goal state is more complex to express if we do not have arithmetic nor aggregate functions that let us count the number of cells coloured with  \texttt{null}. In our instances we define the goal as follows. 
Let $g$ be the maximum allowed number of non-null cells in order to satisfy the goal state. We require that there exist $g$ different cells such that any other cell is \texttt{null}. For instance, requiring at most 2 non-null cells creates the following statement:

{\scriptsize
\begin{verbatim}
    (:goal  ;; at most 2 cells are not null, i.e., g=2
        (exists (?x1 ?x2 ?y1 ?y2 - number) 
            (and
                (or     ;; cell 1 != cell 2. 
                    (not (= ?x1 ?x2))
                    (not (= ?y1 ?y2)))
                (forall (?x3 ?y3 - number) 
                    (or ; Or is one of cell 1 or cell 2, or is null
                        (and (= ?x1 ?x3) (= ?y1 ?y3)) 
                        (and (= ?x2 ?x3) (= ?y2 ?y3))
                        (coloured ?x3 ?y3 null))))))
\end{verbatim}
}

The length of this goal is $\Theta(g^2)$, since the $g$ cells must be pair-wise different.
Again, this would be easier to do with Essence Prime, where we could add an \texttt{atleast} global constraint like the one in the given Essence Prime model. 

Note that the viewpoint of the presented PDDL model is similar to the one presented in the Essence Prime model of Section~\ref{sec:eprime_model}, that is, \texttt{hand} plays a similar role in both cases, and the \texttt{coloured} predicate defines the same information than \texttt{grid}. We leave as future work comparing the given state-oriented Essencie Prime model with an action-oriented Essence Prime model,  which would be similar to an ideally compacted version of the proposed PDDL model.


\section{Empirical Evaluation}

In this section we evaluate the performance of our constraint and PDDL models of Plotting. To facilitate this study, we developed a simple instance generator parameterised on the dimensions of the grid and the number of different colours. It can produce all instances with those parameters, or a random single instance. 
 
For our experimental setup, we use \savilerow~\cite{savilerow} 1.9.0 with CaDiCaL~\cite{cadical} as its backend solver and the Fast Downward~\cite{fastdownward} 20.06+ planner. Experiments are executed in a AMD Opteron\textsuperscript{\textregistered} Processor 6272, and each process was given a limit of 4GB of memory and 1-hour timeout.


As presented in Table \ref{tab:Results}, we have generated squared instances of $n\times n$ cells, for $n$ from 5 to 9, and for each $n$ we consider 3, 4 and 5 colours, going beyond the scenarios proposed by the original game. Moreover, each instance has been replicated with a different goal number of non-empty cells $g$ from 0 to $n^2-1$. Finally, recall from Section~\ref{sec:PlanningAsSatisfiability} that we consider a sequence of decision instances from $1$ to $n^2-g$. Therefore, \emph{count} column shows the number of decision instances obtained with all different goals and all different number of time steps.

The extensive use of quantifiers and the complex conditional effects in the PDDL model are a heavy burden for Fast Downward. The planner is not able to pre-process the PDDL model of sizes greater than 3 within the time-out and therefore none of the considered PDDL instances could be solved.
Table \ref{tab:Results} summarises the obtained results with the \eprime model.
As expected, the more
colours in the problem the harder it becomes. When arriving to 9 by 9 problems,
\savilerow is only able to solve trivial instances and the rest exceed the
allowed memory threshold.
In terms of steps, we are able to see solved instances up to 33 steps for easy
problems. If we consider the increasing sequence of satisfiability questions
for each instance, we observe that in most cases the cost of
pre-processing (i.e. Savile Row) grows linearly, while CaDiCaL exhibits a
classical phase transition around the first satisfiable problem.
\begin{table}[t!]
\centering
\begin{tabular}{|ccc|ccc|cc|}
\hline
n & colours & count & \% solved & SAT & UNSAT & pre-processing & solving \\
\hline
5  & 3   & 325    & 0.98      & 163   & 156  & 37.43        & 28.63             \\
5  & 4   & 325    & 1.00      & 132   & 193  & 37.26        & 127.34            \\
5  & 5   & 325    & 1.00      & 25    & 300  & 44.69        & 0.81              \\
\hline
6  & 3   & 666    & 0.96      & 377   & 264  & 103.35       & 89.74             \\
6  & 4   & 666    & 0.93      & 336   & 281  & 94.81        & 137.07            \\
6  & 5   & 666    & 0.93      & 291   & 331  & 91.13        & 163.92            \\
\hline
7  & 3   & 1225   & 0.68      & 426   & 413  & 141.68       & 132.35            \\
7  & 4   & 1225   & 0.66      & 392   & 417  & 135.88       & 175.60            \\
7  & 5   & 1225   & 0.64      & 263   & 526  & 126.20       & 217.87            \\
\hline
8  & 3   & 2080   & 0.40      & 291   & 550  & 136.58       & 164.88            \\
8  & 4   & 2080   & 0.40      & 251   & 576  & 134.37       & 174.88            \\
8  & 5   & 2080   & 0.41      & 172   & 689  & 141.68       & 224.32            \\
\hline
9  & 3   & 3321   & 0.27      & 158   & 735  & 157.28       & 115.68            \\
9  & 4   & 3321   & 0.27      & 142   & 740  & 155.80       & 148.40            \\
9  & 5   & 3321   & 0.27      & 128   & 758  & 157.94       & 163.23 \\          
\hline
\end{tabular}
\caption{\label{tab:Results}Summarised results of the considered sets of instances using \savilerow. Column \emph{n} represents the width and height of the instances. \emph{pre-processing} and \emph{solving} columns show the mean running time of each step, in seconds.}
\end{table}

%

\section{Conclusions and Further Work}

Using Essence Prime and PDDL, we have presented two models for the tile-matching Plotting video game. Savile Row with CaDiCaL is able to solve much bigger instances than the ones found in the game, while Fast Downward is not able to preprocess non-trivial instances. Both models capture the complex state transitions between steps in the puzzle. Since Essence Prime is a more expressive language, key points in the model are much easier to encode. Native constructs for Essence Prime to express planning-specific primitives would further help into making the encoding of planning problems easier and more natural.

On the other hand, the lack of support of some crucial PDDL features such as multi-valued variables, functional symbols and numeric reasoning makes the modelling of problems with complex transitions a cumbersome and error-prone process. Better support for these would not only motivate the introduction of new problems but more crucially open the possibility of new solving techniques.

\bibliographystyle{splncs04} 
\bibliography{plotting}

\newpage

\begin{subappendices}
\renewcommand{\thesection}{\Alph{section}}%
\section{Essence Prime Model}\label{eprimemodel}

{\scriptsize
\begin{verbatim}
language ESSENCE' 1.0

given initGrid : matrix indexed by[int(1..gridHeight), int(1..gridWidth)] of int(1..)
letting GRIDCOLS be domain int(1..gridWidth)
letting GRIDROWS be domain int(1..gridHeight)
letting NOBLOCKS be gridWidth * gridHeight
letting COLOURS be domain int(1..max(flatten(initGrid)))
letting EMPTY be 0
letting EMPTYANDCOLOURS be domain int(EMPTY) union COLOURS

given goalBlocksRemaining : int(1..NOBLOCKS)

given noSteps : int(1..)
letting STEPSFROM1 be domain int(1..noSteps)
letting STEPSFROM0 be domain int(0..noSteps)

find fpRow : matrix indexed by[STEPSFROM1] of int(0..gridHeight)
find fpCol : matrix indexed by[STEPSFROM1] of int(0..gridWidth)
find grid : matrix indexed by[STEPSFROM0, GRIDROWS, GRIDCOLS]
                           of EMPTYANDCOLOURS
find hand : matrix indexed by[STEPSFROM0] of COLOURS
find wallFall : matrix indexed by[STEPSFROM1] of int(0..gridHeight)

such that

$ Initial state:
forAll gCol : GRIDCOLS .
  forAll gRow : GRIDROWS .
    grid[0, gRow, gCol] = initGrid[gRow, gCol],

$ Goal state:
atleast(flatten(grid[noSteps,..,..]),
        [NOBLOCKS - goalBlocksRemaining],
        [EMPTY]),


$ Each move must do something useful:
forAll step : STEPSFROM1 .
  sum(flatten(grid[step-1,..,..])) > sum(flatten(grid[step,..,..])),

$ Exactly one fp axis must be 0. (exclusive OR, only ONE fired angle)
forAll step : STEPSFROM1 .
  (fpRow[step] * fpCol[step]) = 0 /\ (fpRow[step] + fpCol[step]) > 0,

forAll step : STEPSFROM1 .
  (hand[step-1] = hand[step])
  =
  (
    $ Fired down col, hitting wall
    (
      (forAll colBlock : GRIDROWS .
        ((grid[step-1,colBlock,fpCol[step]] = hand[step-1]) \/
         (grid[step-1,colBlock,fpCol[step]] = EMPTY))
      )
    ) \/
    $ Fired row, hitting wall, dropping through hand-colour only.
    (
      $ along the row
      (
        forAll rowBlock : GRIDCOLS .
          ((grid[step-1,fpRow[step],rowBlock] = hand[step-1]) \/
           (grid[step-1,fpRow[step],rowBlock] = EMPTY))
      ) /\
      $ down the column
      (
       forall rowBeneath : GRIDROWS .
         (rowBeneath > fpRow[step] ->
           ((grid[step-1,rowBeneath,gridWidth] = hand[step-1]) \/
            (grid[step-1,rowBeneath,gridWidth] = EMPTY))
         )
      )
    )
  ),


forAll step : STEPSFROM1 .
  forAll gRow : GRIDROWS .
    forAll gCol : GRIDCOLS .
      (grid[step,gRow,gCol] = EMPTY)
      =
      (
        $ When a cell is EMPTY, it stays EMPTY
        (grid[step-1,gRow,gCol] = EMPTY) \/
        $ Deleted by shot down column
        (
          $ The right column
          (fpCol[step] = gCol) /\
          $ same colour as hand
          (grid[step-1,gRow,gCol] = hand[step-1]) /\
          $ Nothing blocking the way
          (forAll blockAbove : int(1..gRow-1) .
            ((grid[step-1,blockAbove,fpCol[step]] = hand[step-1]) \/
             (grid[step-1,blockAbove,fpCol[step]] = EMPTY))
          )
        ) \/
        $ Deleted by shot along row
        (
          $ The right row
          (fpRow[step] = gRow) /\
          $ same colour as hand
          (grid[step-1,gRow,gCol] = hand[step-1]) /\
          $ no block above
          ((gRow = 1) \/
           (grid[step-1,gRow-1,gCol] = EMPTY)) /\
          $ nothing blocking way
          (forAll blockLeft : int(1..gCol-1) .
            ((grid[step-1,gRow,blockLeft] = hand[step-1]) \/
             (grid[step-1,gRow,blockLeft] = EMPTY))
          )
        ) \/
        $ Deleted by shot along row, then down col
        (
          $ Only the final column
          (gCol = gridWidth) /\
          $ fpRow is above this
          (fpRow[step] < gRow) /\
          $ same colour as hand
          (grid[step-1,gRow,gridWidth] = hand[step-1]) /\
          $ Nothing blocking the way on the row
          (forAll rowBlock : int(1..gridWidth) .
             ((grid[step-1,fpRow[step],rowBlock] = hand[step-1]) \/
              (grid[step-1,fpRow[step],rowBlock] = EMPTY))
          ) /\
          $ Nothing blocking the way on the final column
          (forAll colBlock : int(1..gRow-1) .
             (colBlock > fpRow[step]) ->
             ((grid[step-1,colBlock,gridWidth] = hand[step-1]) \/
              (grid[step-1,colBlock,gridWidth] = EMPTY))
          ) /\
          $ Empty above the firing row on the final column
          (forAll colBlock : int(1..gRow-1) .
             (colBlock < fpRow[step]) ->
             (grid[step-1,colBlock,gridWidth] = EMPTY)
          )
        ) \/
        $ Fall from this cell to become empty - row shot underneath
        (
          $ There was no block above
          ((grid[step-1,gRow-1,gCol] = EMPTY) \/
           (gRow = 1)) /\
          $ Deletion below
          (fpRow[step] > gRow) /\
          (forAll delBlock : int(1..gCol) .
            ((grid[step-1,fpRow[step],delBlock] = hand[step-1]) \/
             (grid[step-1,fpRow[step],delBlock] = EMPTY)))
        ) \/
        $ Final Column shot along a row consuming several blocks underneath
        (
          $ Only the final column
          (gCol = gridWidth) /\
          $ There was a wallfall - this implies a successful row shot.
          (wallFall[step] > 0) /\
          $ The shot was beneath here
          (fpRow[step] > gRow) /\
          $ Nothing there to fall into here
          (grid[step-1,gRow-wallFall[step],gridWidth] = EMPTY \/
           gRow-wallFall[step] < 1)
        )
      ),


forAll step : STEPSFROM1 .
  forAll gRow : GRIDROWS .
    forAll gCol : GRIDCOLS .
      (grid[step,gRow,gCol] = grid[step-1,gRow,gCol])
      =
      (
        $ It was empty
        (grid[step-1,gRow,gCol] = EMPTY) \/
        $ Fired beneath this row, not far enough to cause fall:
        (
          (fpRow[step] > gRow) /\
          (exists blockLeft : int(1..gCol) .
             ((grid[step-1,fpRow[step],blockLeft] != EMPTY) /\
              (grid[step-1,fpRow[step],blockLeft] != hand[step-1]))
          )
        ) \/
        $ Fired along this row, but something in the way
        (
          (fpRow[step] = gRow) /\
          (exists blockLeft : int(1..gCol-1) .
            ((grid[step-1, gRow, blockLeft] != EMPTY) /\
             (grid[step-1, gRow, blockLeft] != hand[step-1]))
          )
        ) \/
        $ Fired along row above, cols except last
        (
          (gCol < gridWidth) /\
          (fpRow[step] != 0) /\
          (fpRow[step] < gRow)
        ) \/
        $ Fired along row above, last col. Sth in way on row or last col.
        (
          (gCol = gridWidth) /\
          (fpRow[step] != 0) /\
          (fpRow[step] < gRow) /\
          (
            (exists rowBlock : int(1..gridWidth) .
             ((grid[step-1, fpRow[step], rowBlock] != EMPTY) /\
              (grid[step-1, fpRow[step], rowBlock] != hand[step-1]))
            ) \/
            (exists colBlock : int(1..gRow-1) .
             ((colBlock >= fpRow[step]) /\
              (grid[step-1, colBlock, gridWidth] != EMPTY) /\
              (grid[step-1, colBlock, gridWidth] != hand[step-1]))
            )
          )
        ) \/
        $ Fired down this column, but something in way
        (
          (fpCol[step] = gCol) /\
          (exists blockAbove : int(1..gRow-1) .
            ((grid[step-1,blockAbove,gCol] != EMPTY) /\
             (grid[step-1,blockAbove,gCol] != hand[step-1])))
        ) \/
        $ Fired down a different column
        (
          (fpCol[step] != 0) /\
          (fpCol[step] != gCol)
        ) \/
        $ This row or below. Same colour block falls here. All but last col.
        (
          (gCol < gridWidth) /\
          (fpRow[step] >= gRow) /\
          (forAll delBlock : int(1..gCol) .
            ((grid[step-1,fpRow[step],delBlock] = hand[step-1]) \/
             (grid[step-1,fpRow[step],delBlock] = EMPTY))) /\
          (grid[step-1,gRow-1,gCol] = grid[step-1,gRow,gCol])
        ) \/
        $ This row or below. Same colour block falls here. Last col.
        (
          (gCol = gridWidth) /\
          (fpRow[step] >= gRow) /\
          (wallFall[step] > 0) /\
          (grid[step-1,gRow-wallFall[step],gCol] = grid[step-1,gRow,gCol])
        )
      ),

forAll step : STEPSFROM1 .
  forAll gRow : GRIDROWS .
    forAll gCol : GRIDCOLS .
      ((grid[step,gRow,gCol] != grid[step-1,gRow,gCol]) /\
       (grid[step,gRow,gCol] != EMPTY))
      =
      (
        $ Fall from above. Not rightmost col.
        (
          (gCol < gridWidth) /\
          $there was a block above
          (grid[step-1,gRow-1,gCol] != EMPTY) /\
          $Deletion here or below
          (fpRow[step] >= gRow) /\
          (forAll delBlock : int(1..gCol) .
            ((grid[step-1,fpRow[step],delBlock] = hand[step-1]) \/
             (grid[step-1,fpRow[step],delBlock] = EMPTY))) /\
          $ Is now the same as the block above.
          (grid[step,gRow,gCol] = grid[step-1,gRow-1,gCol]) /\
          $ Which was a different colour
          (grid[step-1,gRow,gCol] != grid[step-1,gRow-1,gCol])
        ) \/
        $ Fall from above. Rightmost col.
        (
          (gCol = gridWidth) /\
          $ WallFall implies successful row shot
          (wallFall[step] > 0) /\
          $ Shot here or below
          (fpRow[step] >= gRow) /\
          $ Is now the same as the block above
          (grid[step,gRow,gridWidth] =
           grid[step-1,gRow-wallFall[step],gridWidth]) /\
          $ Which was a different colour
          (grid[step-1,gRow,gridWidth] !=
           grid[step-1,gRow-wallFall[step],gridWidth])
        ) \/
        $ Cell swaps with hand: row shot.
        (
          (gRow = fpRow[step]) /\
          $ The row shot
          (forAll colsLeft : int(1..gCol-1) .
            (grid[step-1,fpRow[step],colsLeft] = hand[step-1]) \/
            (grid[step-1,fpRow[step],colsLeft] = EMPTY)) /\
          $ At least one cell has to match the hand
          (exists colsLeft : int(1..gCol-1) .
            (grid[step-1,fpRow[step],colsLeft] = hand[step-1])) /\
          $ Exchanges with the hand
          (hand[step] = grid[step-1,fpRow[step],gCol]) /\
          (hand[step-1] = grid[step,fpRow[step],gCol]) /\
          $ Which was a different colour
          (hand[step-1] != grid[step-1,fpRow[step],gCol])
        ) \/
        $ Cell swaps with hand: col shot
        (
          (gCol = fpCol[step]) /\
          $ The col shot
          (forAll rowsAbove : int(1..gRow-1) .
            (grid[step-1,rowsAbove,fpCol[step]] = hand[step-1]) \/
            (grid[step-1,rowsAbove,fpCol[step]] = EMPTY)) /\
          $ At least one cell has to match the hand
          (exists rowsAbove : int(1..gRow-1) .
            (grid[step-1,rowsAbove,fpCol[step]] = hand[step-1])) /\
          $ Exchanges with the hand
          (hand[step] = grid[step-1,gRow,fpCol[step]]) /\
          (hand[step-1] = grid[step,gRow,fpCol[step]]) /\
          $ Which was a different colour
          (hand[step-1] != grid[step-1,gRow,fpCol[step]])
        ) \/
        $ Cell swaps with hand: row then down last col.
        (
          $ rightmost col
          (gCol = gridWidth) /\
          $ fpRow is above this
          (fpRow[step] < gRow) /\
          $ Nothing blocking the way on the row
          (forAll rowBlock : int(1..gridWidth-1) .
            ((grid[step-1,fpRow[step],rowBlock] = hand[step-1]) \/
             (grid[step-1,fpRow[step],rowBlock] = EMPTY))
          ) /\
          $ Nothing blocking the way on the final column
          (forAll colBlock : int(1..gRow-1) .
            (colBlock >= fpRow[step]) ->
              ((grid[step-1,colBlock,gridWidth] = hand[step-1]) \/
               (grid[step-1,colBlock,gridWidth] = EMPTY))
          ) /\
          $ But there must exist one hand block on firing row or final col.
          ((exists rowBlock : int(1..gridWidth-1) .
              grid[step-1,fpRow[step],rowBlock] = hand[step-1])
           \/
           (exists colBlock : int(1..gRow-1) .
              colBlock >= fpRow[step] /\
              grid[step-1,colBlock,gridWidth] = hand[step-1])
          ) /\
          $ Exchanges with hand
          (hand[step] = grid[step-1,gRow,gridWidth]) /\
          (hand[step-1] = grid[step,gRow,gridWidth]) /\
          $ Which was a different colour
          (hand[step-1] != grid[step-1,gRow,gridWidth])
        )
      ),

forAll step : STEPSFROM1 .
 forAll i : int (1..gridHeight) .
  (wallFall[step] = i)
  =
  (exists row : int(2..gridHeight) .
    (fpRow[step] = row) /\
    $ Consumed row to the rightmost column
    (forAll col : int(1..gridWidth) .
      grid[step-1,row,col] = EMPTY \/
      grid[step-1,row,col] = hand[step-1]) /\
    $ Something to fall
    (grid[step-1,row-1,gridWidth] != EMPTY) /\
    $ Fell this far
    (forAll underRow : int (row..row+i-1) .
      grid[step-1,underRow,gridWidth] = hand[step-1]) /\
    $ And no further
    ((grid[step-1,row+i,gridWidth] != hand[step-1]) \/
     (row+i > gridHeight))
  ),

true      
\end{verbatim}
}

\section{PDDL Model} \label{pddlmodel}

{\scriptsize
\begin{verbatim}
(define (domain plotting)
    (:requirements :typing :equality :universal-preconditions :conditional-effects)
    (:types number colour)
    (:constants null wildcard - colour)

    (:predicates
        (hand ?c - colour)
        (coloured ?r ?c - number ?c - colour)
        (succ ?n1 ?n2 - number)
        (gt ?n1 ?n2 - number)
        (pred ?n1 ?n2 - number)
        (lt ?n1 ?n2 - number)
        (distance ?n1 ?n2 ?n3 - number)

        (isfirstcolumn ?n - number) ;; is the first column?
        (islastcolumn ?n - number) ;; is the last column?
        (istoprow ?n - number) ;; is the top row?
        (isbottomrow ?n - number) ;; is the bottom column?

        ;; are we allowed to fire on those?
        (blockedcol ?n - number)
        (blockedrow ?n - number)
    )

    ;; removing a partial row:
    ;;    - we are removing part of a row and there is a next block that has a different
    ;;      colour than the first one.
    (:action shoot-partial-row
        :parameters (?r - number ?t - number ?c - colour)
        :precondition
            (and
                ;; there exists a number that is the successor of the to
                ;; and it has a different colour than c
                (exists (?col - number)
                    (and
                        (succ ?col ?t)
                        (not (coloured ?r ?col ?c))
                        (not (coloured ?r ?col null))))
                      
                ;; exist some column before ?t or ?t that has colour ?c      
                (exists (?col - number)
                    (and
                        (or (lt ?col ?t) (=  ?col ?t))
                        (coloured ?r ?col ?c)))

                ;; stop possible weird stuff
                (not (= ?c null))
                (not (= ?c wildcard))
                ;; colour block and hand is the same (we avoid null movements)
                (or (hand ?c) (hand wildcard))

                ;; from the start, all the blocks up to ?t have either the colour ?c or are null
                (forall (?col - number) 
                    ;; For all columns, either:
                    (or
                        ;; we are out of bounds or
                        (gt ?col ?t)
                 
                        ;; the middle colors and color of ?t are null or the correct one
                        (coloured ?r ?col ?c)
                        (coloured ?r ?col null))))
        :effect
            (and
                ;; Change hands colour and
                ;; The next cell that we cannot remove gets the colour from the hand
                (forall (?nextcolumn - number ?nextcolour - colour)
                    (when 
                        (and
                            ;; there is a new row down this one
                            (succ ?nextcolumn ?t) 
                            ;; and the cell is coloured ?nextcolour
                            (coloured ?r ?nextcolumn ?nextcolour))
                        ;; The hand gets a new colour
                        (and
                            ;; change next cell colour
                            (not (coloured ?r ?nextcolumn ?nextcolour))
                            (coloured ?r ?nextcolumn ?c)
                            ;; change hand colours
                            (hand ?nextcolour)
                            (not (hand ?c))
                            (not (hand wildcard)))))

                ;; finally move everything downwards. we have 2 cases:
                ;;  - we are on the top row
                ;;  - we are on another row
                (forall (?currentrow ?nextrow ?currentcol - number)
                    (and
                        ;; We are on the top row: we must restore the "null" colour
                        (forall (?currentcolor - colour)
                            (when
                                (and 
                                    ;; we are on the top row 
                                    (istoprow ?currentrow)
                                    ;; The column is in the correct range
                                    (or (lt ?currentcol ?t) (= ?currentcol ?t))
                                    ;; We identify the colour of the cell
                                    (coloured ?currentrow ?currentcol ?currentcolor)
                                    ;; avoid contradiction
                                    (not (coloured ?currentrow ?currentcol null)))
                                (and
                                    (not (coloured ?currentrow ?currentcol ?currentcolor))
                                    (coloured ?currentrow ?currentcol null))))

                        ;; We are on any other row: disable the current colour and change the next colour
                        (forall (?currentcolor ?nextcolor - colour)
                            (when
                                ;; when the row is on top of the one we deleted, we "decrease"
                                ;; one position downwards 
                                (and
                                    (lt ?currentrow ?r)            
                                    (succ ?nextrow ?currentrow)    
                                    ;; The current column is in the correct range (less than the ?t)
                                    (or (lt ?currentcol ?t) (= ?currentcol ?t))

                                    ;; here we ensure that the cells have the pertaining colours
                                    (coloured ?currentrow ?currentcol ?currentcolor)
                                    (coloured ?nextrow ?currentcol ?nextcolor)
                                    ;; avoid the contradiction in the effect if both colours are equal
                                    (not (= ?currentcolor ?nextcolor)))
                                (and ;; and as an effect we change the lower row
                                    (not (coloured ?nextrow ?currentcol ?nextcolor))
                                    (coloured ?nextrow ?currentcol ?currentcolor))))))))

    (:action shoot-column
        :parameters (?column - number ?t - number ?c - colour)
        :precondition 
            (and
                ;; the successor of the to is of a different colour
                ;; or we are eating the whole column.
                ;; note we don't have to check null colour because gravity
                (or
                    (exists (?row - number)
                        (and
                            (succ ?row ?t)
                            (not (coloured ?row ?column ?c))))
                    (isbottomrow ?t))

                ;; stop possible weird stuff
                (not (= ?c null))
                (not (= ?c wildcard))
                ;; colour block and hand is the same (we avoid null movements)
                (or (hand ?c) (hand wildcard))

                ;; all the middle blocks also have the same colour
                (forall (?row - number) 
                    (or
                        ;; either we are out of bounds or
                        (gt ?row ?t)
                        ;; we are exactly in ?t and it has the correct colour
                        (and (= ?row ?t) (coloured ?row ?column ?c))
                        ;; we are on top of ?t and therefore it can be either ?c or null
                        (and
                            (lt ?row ?t)
                            (or
                                (coloured ?row ?column ?c)
                                (coloured ?row ?column null))))))
        :effect
            (and
                (forall (?runningrow - number)
                    (and
                        ;; we remove the colour of the cells before the "to" (?t)
                        (when 
                            (and
                                (coloured ?runningrow ?column ?c)
                                (or (lt ?runningrow ?t)
                                    (= ?runningrow ?t))
                                (not (coloured ?runningrow ?column null)))
                            (and
                                (coloured ?runningrow ?column null)
                                (not (coloured ?runningrow ?column ?c))))

                        ;; we set the next block colour to the removed cells colour
                        ;; and we set the hands colour to the correct one
                        (forall (?nextcolour - colour)
                            (when
                                (and
                                    (succ ?runningrow ?t)
                                    (coloured ?runningrow ?column ?nextcolour))
                                (and
                                    ;; next cell colour
                                    (not (coloured ?runningrow ?column ?nextcolour))
                                    (coloured ?runningrow ?column ?c)
                                    ;; hand colour
                                    (not (hand wildcard))
                                    (not (hand ?c))
                                    (hand ?nextcolour))))))

                ;; When we are eating the whole column, we set the correct colour onto the hand
                (when
                    (isbottomrow ?t)
                    (and
                        (not (hand wildcard))
                        (hand ?c)))))

    (:action shoot-row-and-column
            :parameters (?r - number ?t - number ?c - colour)
            :precondition
                (and
                    ;; rows of the shot are coherent
                    (gt ?t ?r)

                    ;; stop possible weird stuff
                    (not (= ?c null))
                    (not (= ?c wildcard))
                    ;; colour block and hand is the same (we avoid null movements)
                    (or (hand ?c) (hand wildcard))

                    ;; all the blocks in the row and in the corresponding part of the
                    ;; column are either the same colour or are empty
                    (forall (?row ?col - number) 
                        (and
                            (imply
                                (= ?r ?row)
                                (or
                                    (coloured ?r ?col ?c)
                                    (coloured ?r ?col null)))
                            (imply
                                (and
                                    (gt ?row ?r)
                                    (not (gt ?row ?t))
                                    (islastcolumn ?col))
                                (or
                                    (coloured ?row ?col ?c)
                                    (coloured ?row ?col null)))))

                    ;; we have to eat at least one of those!
                    (exists (?row ?col - number) 
                        (or 
                            (and
                                (= ?r ?row)
                                (coloured ?r ?col ?c))
                        
                            (and
                                (gt ?row ?r)
                                (not (gt ?row ?t))
                                (islastcolumn ?col)
                                (coloured ?row ?col ?c))))
                            
                    ;; either ?t is the last row, or in its successor there is a different colour
                    (or
                        (isbottomrow ?t)
                        (exists (?nextrow ?lastcol - number)
                            (and
                                (succ ?nextrow ?t)
                                (islastcolumn ?lastcol)
                                (not (coloured ?nextrow ?lastcol ?c))
                                (not (coloured ?nextrow ?lastcol null))))))
            :effect
                (and
                    ;; move everything downwards except the last column.
                    ;; 2 cases: we are on the top row or  we are on a middle row
                    (forall (?currentrow ?currentcol ?nextrow - number)
                        (and
                            ;; case 1 - We are on the top row: we must restore the "null" colour
                            (when
                                (istoprow ?currentrow)
                                (and
                                    (not (coloured ?currentrow ?currentcol ?c))
                                    (coloured ?currentrow ?currentcol null)))

                            ;; case 2 - We are on the middle row: disable the current colour and change the next colour
                            (forall (?currentcolor ?nextcolor - colour)
                                (when
                                    (and
                                        (not (istoprow ?nextrow))        ;; is not top row
                                        (not (islastcolumn ?currentcol)) ;; is not last column
                                        ;; position the "pointers"
                                        (lt ?currentrow ?r)              
                                        (succ ?nextrow ?currentrow)
                                        ;; ensure that the cells have the pertaining colours
                                        (coloured ?currentrow ?currentcol ?currentcolor)
                                        (coloured ?nextrow ?currentcol ?nextcolor)
                                        ;;avoid effect contradiction (if both colours are equal)
                                        (not (= ?currentcolor ?nextcolor))
                                    )
                                    (and ;; and as an effect we change the lower row
                                        (not (coloured ?nextrow ?currentcol ?nextcolor))
                                        (coloured ?nextrow ?currentcol ?currentcolor))))))

                    ;; The waterfall effect
                    (forall (?currentrow ?nextrow ?lastcolumn ?d ?dplus1 ?d2 - number ?nextcolour ?currentcolour - colour)
                        (and

                            ;; unconditionally, if we reach the end of any row at all, the cell
                            ;; on the top right of the grid will get a null.
                            (when
                                (and
                                    (istoprow ?currentrow)
                                    (islastcolumn ?lastcolumn)
                                    (coloured ?currentrow ?lastcolumn ?nextcolour)
                                    (not (coloured ?currentrow ?lastcolumn null)))
                                (and
                                    (not (coloured ?currentrow ?lastcolumn ?nextcolour))
                                    (coloured ?currentrow ?lastcolumn null)))

                   
                            ;; we act only on the set of cells between ?r and ?t
                            ;; base case: we are on top of the row
                            (when 
                                (and
                                    ;; we consider the case of the last column
                                    (islastcolumn ?lastcolumn)
                                    ;; distance between the row we shoot at and where we stop is ?d
                                    (distance ?r ?t ?d)
                                    ;; and we calculate the distance we have to move cells downwards
                                    (succ ?dplus1 ?d)
                                    ;; we need to fix this here to be able to compute the distance with the
                                    ;; nextrow pointer, which is the one we use to change 
                                    (istoprow ?currentrow)
                                    ;; we are on top of the ?t
                                    (or (lt ?nextrow ?t) (= ?nextrow ?t))
                                    ;; ?d2 is the distance between nextrow currentrow
                                    ;; and this distance is less than what we should copy
                                    (distance ?currentrow ?nextrow ?d2)
                                    (lt ?d2 ?dplus1)

                                    ;; and the colours are correct and not null
                                    (coloured ?nextrow ?lastcolumn ?nextcolour)
                                    (not (= ?nextcolour null)))
                                (and ;; we switch colours
                                    (not (coloured ?nextrow ?lastcolumn ?nextcolour))
                                    (coloured ?nextrow ?lastcolumn null)))
                            ;; other cases
                            (when 
                                (and
                                     ;; we consider the case of the last column
                                    (islastcolumn ?lastcolumn)
                                    ;; distance between the row we shoot at and where we stop is ?d
                                    (distance ?r ?t ?d)
                                    ;; and we calculate the distance we have to move cells downwards
                                    (succ ?dplus1 ?d)
                                    ;; we are on top of the ?t
                                    (or (lt ?nextrow ?t) (= ?nextrow ?t))
                                    ;; and the distance is ?dplus1
                                    (distance ?nextrow ?currentrow ?dplus1)
                                    ;; and the colours are correct and different
                                    (coloured ?currentrow ?lastcolumn ?currentcolour)
                                    (coloured ?nextrow ?lastcolumn ?nextcolour)
                                    (not (= ?currentcolour ?nextcolour)))
                                (and ;; we switch colours
                                    (not (coloured ?nextrow ?lastcolumn ?nextcolour))
                                    (coloured ?nextrow ?lastcolumn ?currentcolour)))
                            ;; finally, we change the colour of the cell on the bottom of the ?t.
                            (when 
                                (and
                                     ;; we consider the case of the last column
                                    (islastcolumn ?lastcolumn)
                                    ;; distance between the row we shoot at and where we stop is ?d
                                    (distance ?r ?t ?d)
                                    ;; and we calculate the distance we have to move cells downwards
                                    (succ ?dplus1 ?d)
                                    (succ ?nextrow ?t)
                                    ;; and the colour is correct and different from null
                                    (coloured ?nextrow ?lastcolumn ?nextcolour))
                                (and ;; we switch colours
                                    (not (coloured ?nextrow ?lastcolumn ?nextcolour))
                                    (coloured ?nextrow ?lastcolumn ?c)))))
                   
                   ;; Change hands colour:
                   ;; base case when we eat all the last column or is all null
                   (when 
                        (isbottomrow ?t)
                        (and 
                            (hand ?c)
                            (not (hand wildcard))))

                    ;; - we are not the last row before ground, therefore
                    ;; the colour of the hand becomes the one under the last cell of the row
                    (forall (?nextrow ?lastcol - number ?nextcolour - colour)
                        (and 
                            (when 
                                (and
                                    ;; there is another row down this one
                                    (succ ?nextrow ?t)
                                    ;; and we are on the last column
                                    (islastcolumn ?lastcol)
                                    ;; and the next (down) cell is coloured ?nextcolour
                                    (coloured ?nextrow ?lastcol ?nextcolour))
                                (and
                                    ;; we change the colour of the next (down) cell
                                    (not (coloured ?nextrow ?lastcol ?nextcolour))
                                    (coloured ?nextrow ?lastcol ?c)
                                    ;; we set the hand to the next cell colour
                                    (hand ?nextcolour)
                                    (not (hand ?c))
                                    ;; we ensure we lose the wildcard if we had it
                                    (not (hand wildcard))))))))



    (:action shoot-only-full-row
            ;; ?r - what complete row we are shooting.
            ;; ?c - the colour of the range
            :parameters (?r - number ?c - colour)
            :precondition
                (and
                    ;; stop possible weird stuff
                    (not (= ?c null))
                    (not (= ?c wildcard))
                    ;; colour block and hand is the same (we avoid null movements)
                    (or (hand ?c) (hand wildcard))

                    ;; all the blocks in the row and in the corresponding part of the
                    ;; column are either the same colour or are empty
                    (forall (?col - number) 
                        (or
                            (coloured ?r ?col ?c)
                            (coloured ?r ?col null)))

                    ;; we have to eat at least one of those!
                    (exists (?col - number) 
                        (coloured ?r ?col ?c))
                            
                    ;; either ?r is the last row, or in the cell below the last one 
                    ;; we have a different colour than ?c
                    (or
                        (isbottomrow ?r)
                        (exists (?nextrow ?lastcol - number)
                            (and
                                (succ ?nextrow ?r)
                                (islastcolumn ?lastcol)
                                (not (coloured ?nextrow ?lastcol ?c))
                                (not (coloured ?nextrow ?lastcol null))))))
            :effect
                (and
                    ;; move everything downwards
                    ;; 2 cases: we are on the top row or  we are on a middle row
                    (forall (?currentrow ?currentcol ?nextrow - number)
                        (and
                            ;; case 1 - We are on the top row: we must restore the "null" colour
                            (when
                                (istoprow ?currentrow)
                                (and
                                    (not (coloured ?currentrow ?currentcol ?c))
                                    (coloured ?currentrow ?currentcol null)))

                            ;; case 2 - We are on the middle row: disable the current colour and change the next colour
                            (forall (?currentcolor ?nextcolor - colour)
                                (when
                                    (and
                                        (not (istoprow ?nextrow))        ;; is not top row
                                        ;; position the "pointers"
                                        (lt ?currentrow ?r)              
                                        (succ ?nextrow ?currentrow)
                                        ;; ensure that the cells have the pertaining colours
                                        (coloured ?currentrow ?currentcol ?currentcolor)
                                        (coloured ?nextrow ?currentcol ?nextcolor)
                                        ;;avoid effect contradiction (if both colours are equal)
                                        (not (= ?currentcolor ?nextcolor))
                                    )
                                    (and ;; and as an effect we change the lower row
                                        (not (coloured ?nextrow ?currentcol ?nextcolor))
                                        (coloured ?nextrow ?currentcol ?currentcolor))))))

                   ;; Change hands colour:
                   ;; base case when we eat all the last column or is all null
                   (when 
                        (isbottomrow ?r)
                        (and 
                            (hand ?c)
                            (not (hand wildcard))))

                    ;; - we are not the last row before ground, therefore
                    ;; the colour of the hand becomes the one under the last cell of the row
                    (forall (?nextrow ?lastcol - number ?nextcolour - colour)
                        (and 
                            (when 
                                (and
                                    ;; there is another row down this one
                                    (succ ?nextrow ?r)
                                    ;; and we are on the last column
                                    (islastcolumn ?lastcol)
                                    ;; and the next (down) cell is coloured ?nextcolour
                                    (coloured ?nextrow ?lastcol ?nextcolour))
                                (and
                                    ;; we change the colour of the next (down) cell
                                    (not (coloured ?nextrow ?lastcol ?nextcolour))
                                    (coloured ?nextrow ?lastcol ?c)
                                    ;; we set the hand to the next cell colour
                                    (hand ?nextcolour)
                                    (not (hand ?c))
                                    ;; we ensure we lose the wildcard if we had it
                                    (not (hand wildcard))))))))
)
\end{verbatim}
}
\end{subappendices}

\end{document}